\documentclass[12pt]{article}

\usepackage{scicite}

\usepackage{times}

\usepackage{graphicx}

\usepackage{hyperref}

\newenvironment{sciabstract}{%
\begin{quote} \bf}
{\end{quote}}

\title{\textbf{Title: } Smartphone Camera Oximetry in an Induced Hypoxemia Study} 

\author
{\textbf{Authors: } Jason S. Hoffman,$^{1 \ast \dag}$ Varun Viswanath,$^{2,3 \dag}$  Xinyi Ding,$^{4}$\\ Matthew J. Thompson,$^{5}$ Eric C. Larson,$^{4}$ Shwetak N. Patel,$^{1,6}$ Edward Wang$^{2,3}$\\
\\
\textbf{Affiliations: }\\
\normalsize{$^{1}$Paul G. Allen School of Computer Science and Engineering, University of Washington}\\
\normalsize{$^{2}$Department of Electrical and Computer Engineering, University of California San Diego}\\
\normalsize{$^{3}$The Design Lab, University of California San Diego}\\
\normalsize{$^{4}$Department of Computer Science, Southern Methodist University}\\
\normalsize{$^{5}$Department of Family Medicine, University of Washington}\\
\normalsize{$^{6}$Department of Electrical and Computer Engineering, University of Washington}\\
\\
\normalsize{$^\ast$To whom correspondence should be addressed; E-mail: jasonhof@cs.washington.edu.}\\
\normalsize{$^\dag$These authors contributed equally to this work.}
}

\date{}

\begin{document} 

\baselineskip16pt

\maketitle 

\section*{One Sentence Summary}
Smartphone-based SpO\textsubscript{2} sensing for hypoxemia screening is demonstrated with 81\% sensitivity and 79\% specificity by applying deep learning to a new dataset gathered from test subjects exhibiting the full range of clinically relevant SpO\textsubscript{2} values (70\%-100\%) in a varied FiO\textsubscript{2} study.

\newpage
\section*{Abstract}
\begin{sciabstract}
  Hypoxemia, a medical condition that occurs when the blood is not carrying enough oxygen to adequately supply the tissues, is a leading indicator for dangerous complications of respiratory diseases like asthma, COPD, and COVID-19. While purpose-built pulse oximeters can provide accurate blood-oxygen saturation (SpO\textsubscript{2}) readings that allow for diagnosis of hypoxemia, enabling this capability in unmodified smartphone cameras via a software update could give more people access to important information about their health, as well as improve physicians' ability to remotely diagnose and treat respiratory conditions. In this work, we take a step towards this goal by performing the first clinical development validation on a smartphone-based SpO\textsubscript{2} sensing system using a varied  fraction of inspired oxygen (FiO\textsubscript{2}) protocol, creating a clinically relevant validation dataset for solely smartphone-based methods on a wide range of SpO\textsubscript{2} values (70\%-100\%) for the first time. This contrasts with previous studies, which evaluated performance on a far smaller range (85\%-100\%). We build a deep learning model using this data to demonstrate accurate reporting of SpO\textsubscript{2} level with an overall MAE=5.00\% SpO\textsubscript{2} and identifying positive cases of low SpO\textsubscript{2}$<$90\% with 81\% sensitivity and 79\% specificity. We ground our analysis with a summary of recent literature in smartphone-based SpO\textsubscript{2} monitoring, and we provide the data from the FiO\textsubscript{2} study in open-source format, so that others may build on this work.
\end{sciabstract}

\newpage
\section*{Introduction}

\begin{figure}
    \centering
    \includegraphics[width=16cm]{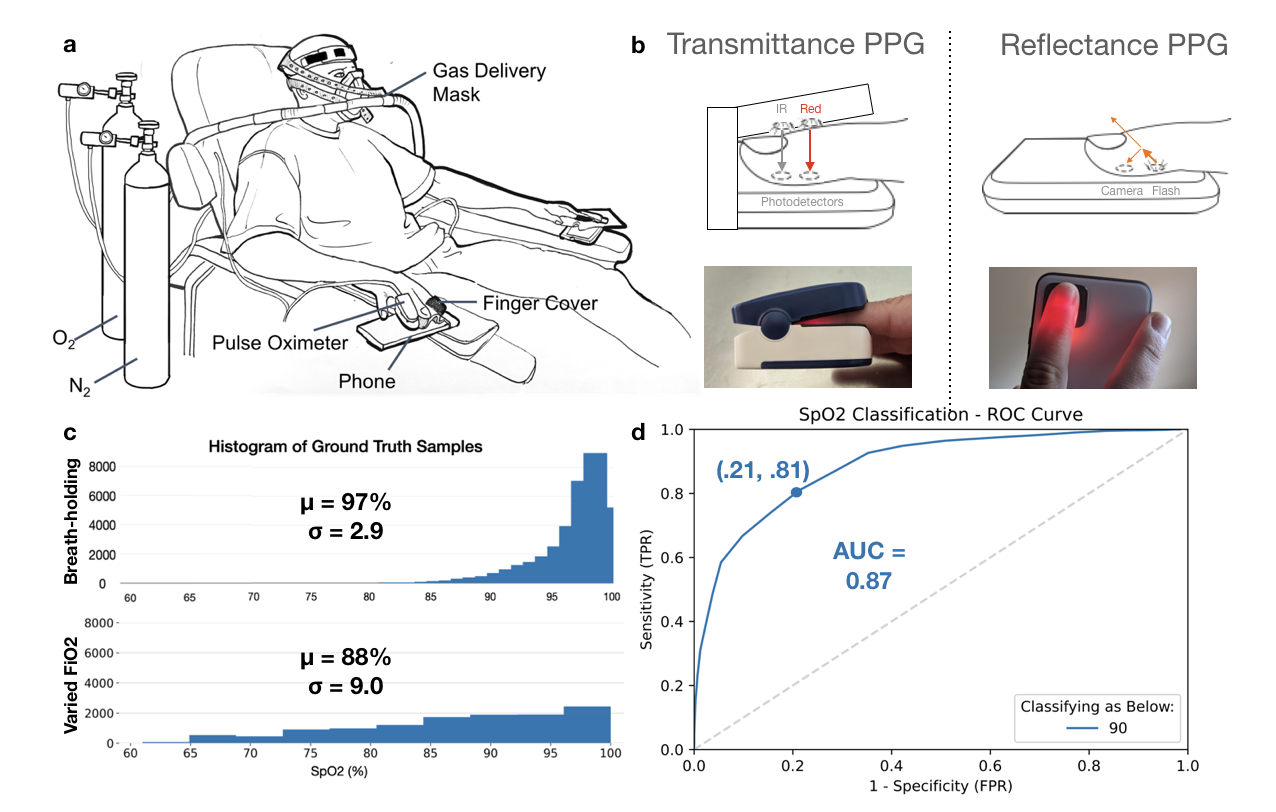}
    \caption{\textbf{A smartphone was used to collect data during a varied FiO\textsubscript{2} experiment.} 
    This experiment produced a more balanced dataset for training our deep learning model than prior work on breath-holding experiments. \textbf{a} Drawn figure of the experimental setup of the varied FiO\textsubscript{2} experiment conducted for this study. The subject breathes a controlled mixture of oxygen and nitrogen to slowly lower the SpO\textsubscript{2} level over a period of 13-19 minutes.
    \textbf{b} Light response was recorded from two fingers on each hand.  One finger was placed over a smartphone camera with flash on to record light response via Reflectance PPG, while a second finger was placed in the fingerclip of a transfer standard pulse oximeter, which emits Red and IR light reports SpO\textsubscript{2} via Transmittance PPG. 
    \textbf{c} Histogram of the data distribution of ground truth of samples from a breath-holding study dataset, adapted from Ding et al. \cite{ding2018measuring} and the histogram of the ground truth distribution from our varied FiO\textsubscript{2} experiment dataset. Our dataset contains more than 1000 samples in both ranges 65\%-80\% and 80\%-90\% SpO\textsubscript{2}, while the example breath-holding study has fewer than 1000 samples total below 90\% SpO\textsubscript{2} \cite{ding2018measuring}. This allows our deep learning model to train and evaluate on the full range of clinically relevant SpO\textsubscript{2} values \cite{us2013pulse}. 
    \textbf{d} Classification results for the smartphone method reveal that $79\%$ of cases of hypoxemia (defined as a low SpO\textsubscript{2} below 90\%) were detected using this method.}
    \label{STM-summ}
\end{figure}

Monitoring blood-oxygen saturation (SpO\textsubscript{2}) with a smartphone, if enabled in an accurate and unobtrusive manner, has the potential to improve health outcomes for those with respiratory illnesses by enabling access to rapid risk assessment outside of face-to-face clinical settings \cite{steinhubl2015emerging}. Recent work on smartphone-based SpO\textsubscript{2} monitors show that these devices may offer the ubiquity and precision necessary to increase access to detection and monitoring of respiratory diseases \cite{bui2020smartphone,ding2018measuring}. This work builds upon these prior findings by being the first to systematically compare smartphone-based SpO\textsubscript{2} monitoring to standalone pulse oximeters on a wide range of clinically-relevant SpO\textsubscript{2} values ($70\% \leq SpO\textsubscript{2} < 100\%$). We show the promise of a smartphone-based system for monitoring SpO\textsubscript{2} by training and testing a deep learning model with data gathered using an unmodified smartphone camera and flash in a varied Fractional Inspired Oxygen (FiO\textsubscript{2}) study (Fig. \ref{STM-summ}a), in which test subjects exhibit SpO\textsubscript{2} levels in the clinically relevant range of 70\%-100\% \cite{us2013pulse}.

Blood-oxygen saturation, reported as SpO\textsubscript{2} percentage, is a clinical measure that informs a physician of the ability of the body to distribute oxygen by revealing the proportion of hemoglobin in the blood currently carrying oxygen.  While baseline SpO\textsubscript{2} level varies slightly (typically $96\%-98\%$ at sea level in otherwise healthy individuals), major deviations from these levels can be a sign of more serious cardiopulmonary disease.
Respiratory illnesses, such as asthma, chronic obstructive pulmonary disease (COPD), pneumonia, and COVID-19, can cause significant decreases in SpO\textsubscript{2}, hypoxemia (low blood oxygen), and potentially hypoxia (low tissue oxygen). Hypoxia can lead to serious complications, such as organ damage to vital organs like the brain or kidneys, and even death, if uncorrected or occurring acutely
for an extended period of time \cite{bickler2017effects}.
Repeated measurements of SpO\textsubscript{2} can be used to assess the severity of a wide range of cardiopulmonary conditions such as asthma and COPD \cite{siddiqui2018severity} 
and detect presence of other illnesses including Idiopathic Pulmonary Fibrosis, Congestive Heart Failure, Diabetic Ketoacidosis, and pulmonary embolism \cite{wilson2012diagnosis,kline2003use,zisman2007prediction}.
90\% SpO\textsubscript{2} has been cited as a threshold below which in-hospital mortality rates increase in COVID-19 patients \cite{xie2020association} and treatment adjustment is needed for primary care patients \cite{jones2003feasibility}.

Blood-oxygen saturation, or SaO\textsubscript{2}, can be directly measured from samples of arterial blood by analyzing the blood samples using a
Arterial Blood Gas (ABG) analysis device, 
which reports the ratio of oxygenated to deoxygenated hemoglobin. 
However, obtaining and analyzing arterial blood samples is invasive and can be technically difficult; therefore, it is typically limited to intensive care or emergency cases.  As a result, clinicians typically rely on the convenience of widely available noninvasive measures of SpO\textsubscript{2} using FDA-cleared, purpose-built devices called pulse oximeters consisting of a finger clip and readout screen (Fig. \ref{STM-summ}b). This device allows physicians to noninvasively monitor SpO\textsubscript{2} for single (spot-check) measures or continuous measures, if necessary to detect changes in status over time.  Some patients, such as those with COPD, use pulse oximeters in home settings to monitor the need for oxygen therapy.
Pulse oximeters typically perform oxygenation measurement via transmittance photoplethysmography (PPG) sensing at the finger tip, clamping around the end of the finger and transmitting red and IR light via LEDs \cite{welch1990pulse}. By measuring the resultant ratio of light transmittance, the devices estimate the absorption properties of the blood, using calibrated curves based on the Beer-Lambert Law to infer blood composition \cite{bui2020smartphone}.  While purpose-built pulse oximetry is non-invasive and accurate across a full range of clinically relevant SpO\textsubscript{2} levels and skin tones, it requires a standalone device.
This reduces access to 
SpO\textsubscript{2} measurements, particularly among patients at home, or by health care workers in lower or middle income countries.
This gap in access to SpO\textsubscript{2} measurements has become more prominent during the COVID-19 pandemic, where home (or out of hospital) monitoring of SpO\textsubscript{2} levels has become a valuable tool in determining the need for clinical care, yet is limited by lack of widely available pulse oximeters
\cite{teo2020early}.

Smartphone-based SpO\textsubscript{2} monitors, especially those that rely only on built-in hardware with no modifications, present an opportunity to detect and monitor respiratory conditions in contexts where pulse oximeters are less available.  Smartphones are widely owned because of their multi-purpose utility, and contain increasingly powerful sensors, including a camera with a LED flash \cite{steinhubl2015emerging,gambhir2018toward,topol2019decade}. Due to their ubiquity, smartphones have been proposed as a decision support tools, indicating the need for health care consultation
\cite{kanakasabapathy2017automated,laksanasopin2015smartphone}. Researchers have used sensors in off-the-shelf smartphone devices to assess many physiological conditions, including detecting voice disorders \cite{mehta2012mobile}, tracking pulmonary function \cite{mehta2012mobile,larson2012spirosmart}, assessing infertility \cite{kanakasabapathy2017automated}, measuring hemoglobin concentration \cite{mannino2018smartphone,wang2017noninvasive}, and estimating changes in blood pressure \cite{wang2018seismo,chandrasekaran2012cuffless}. Smartphone-based solutions for monitoring blood oxygen saturation have been explored, employing various solutions used to gather and stabilize the PPG signal, augment the IR-filtered broad-band camera sensor, and filter the resultant signal for noise or outlier correction. Some solutions require extra hardware, such as a color filter or external light source \cite{mendelson1988noninvasive,carni2016setting,karlen2013detection,tayfur2019reliability,scully2011physiological,bui2020smartphone}, whereas others rely only on the in-built smartphone hardware and employ software techniques to process the PPG signal \cite{tomlinson2018accuracy,nemcova2020monitoring,ding2018measuring,lamonaca2015blood}. Various statistical methods have been used to interpret the results to achieve reasonable accuracy, including the ratio-of-ratios method used by standalone pulse oximeters \cite{bui2020smartphone} and deep learning \cite{ding2018measuring}.  These prior works illustrate the potential for smartphone-based SpO\textsubscript{2} monitors to fill the gaps identified above, but lack validation data on a full range of clinically relevant SpO\textsubscript{2} levels. Prior evaluation techniques for these smartphone-based studies have been limited to a minimum of 80\% SpO\textsubscript{2} using techniques such as breath-holding, which is limited by very short durations of data collection due to participant discomfort, 
limiting the clinical applicability of the findings. The US Food and Drug Administration (FDA) recommends cleared pulse oximeter devices to achieve $<3\%$ accuracy across the full range of clinically relevant data of 70\%-100\% \cite{luks2020pulse,us2013pulse}.  To our knowledge, no prior works have thus far evaluated smartphone-based pulse oximetry on this range of SpO\textsubscript{2} data. 

In order to demonstrate the accuracy of smartphone-based pulse oximetry on the full range of clinically relevant SpO\textsubscript{2} data, we develop and evaluate our system using a varied FiO\textsubscript{2} protocol. This protocol, which is commonly used to validate devices in development towards FDA clearance, requires the test subject to breathe in a combination of oxygen and nitrogen to slowly and safely lower their SpO\textsubscript{2} level to below 65\%. During this test, we record simultaneous video data using an unmodified smartphone camera and SpO\textsubscript{2} "ground truth" reference data from a standard standalone finger clip pulse oximeter, known as a reference standard pulse oximeter \cite{clinimark2010pulse}. In this way, we are able to build a labeled training data set to evaluate the performance of our smartphone-based SpO\textsubscript{2} measurement system across a wide range of clinically relevant levels, and report those here for the first time. Our analysis on 6 subjects reveals that a convolutional neural network (CNN) model is able to achieve a Mean Average Error (MAE) of $5.00\%$ SpO\textsubscript{2} in predicting a new subject's SpO\textsubscript{2} level, after it has been trained only on other subjects' labeled data. To assess potential hypoxemia screening capability, we show that this corresponds to an average sensitivity and specificity of 81\% and 79\% respectively in classifying a new subject's SpO\textsubscript{2} as below 90\%. In addition, when evaluated only on data above 85\% SpO\textsubscript{2}, our results show similar MAE to prior work on smartphone-based SpO\textsubscript{2} sensing.

Our contributions with this study are three-fold: (1) a software application and associated deep learning model for unmodified smartphones that can report SpO\textsubscript{2} measurements with accuracy nearing that of standalone pulse oximeters across a clinically relevant range, (2) analysis of model performance of unmodified smartphone camera oximetry as a screening tool for hypoxemia, and (3) a novel, open-source dataset of a varied FiO\textsubscript{2} experiment captured using an unmodified smartphone camera oximetry system, containing  more than 10,000 labeled SpO\textsubscript{2} sample readings from 6 subjects in the range of 61\%-100\%.

\section*{Results}

\begin{figure}[ht!]
\includegraphics[width=16cm]{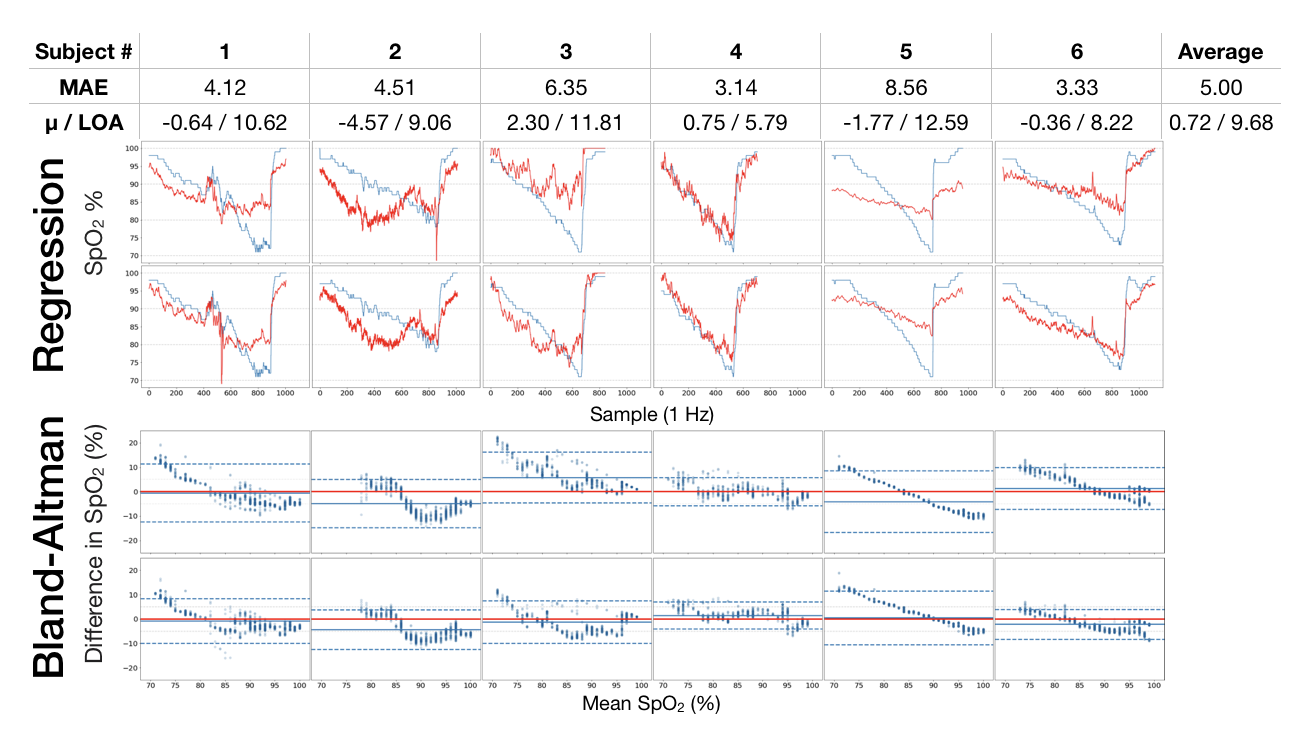}
\caption{\textbf{Regression results displayed as direct performance analysis and Bland-Altman comparison.} MAE averages to 5.00\% over all 6 subjects in the study.  The average difference ($\mu$) and limits of agreement (LOA) average to 0.72 and 9.68, which compare favorably to standalone pulse oximetry devices. \textbf{Table:} MAE and Bland-Altman statistics for CNN evaluation by LOOCV for each subject (n=6) in the study. \textbf{Regression:} Plots of direct performance analysis of regression results.  Model predictions (in red) and ground truth readings (in blue) for the 6 subjects in the FiO\textsubscript{2} study plotted against time of study.  Left hand is on top and right hand is on bottom. \textbf{Bland-Altman:} Bland-Altman plots displaying the spread of predictions against ground truth, revealing that the standalone pulse oximeter and smartphone model perform similarly for most test subjects, with exceptions
discussed in Discussion.}
\label{reg-ba}
\end{figure}

\paragraph*{SpO\textsubscript{2} prediction performance}
Our convolutional neural network (CNN) achieved a MAE of $5.00\%$ SpO\textsubscript{2} when trained and evaluated via leave-one-out cross validation (LOOCV) on the data from the varied FiO\textsubscript{2} study. MAE represents the difference between our model's prediction and the simultaneous reading of the transfer standard pulse oximeter as the ground truth in this study.  Fig. \ref{reg-ba} shows the regression and difference analysis, using Bland-Altman analysis to evaluate the relative accuracy of the smartphone sensing system against the reference standard pulse oximeter. The best within-subject performance is
a MAE of 3.14\%, a mean difference ($\mu$) and Limit of Agreement (LOA) of 0.75\% and 5.79\% (Subject 4). 
The worst within-subject performance is a MAE of 8.56\%, a $\mu$ and LOA of -1.77\% and 12.59\% (Subject 5).  The reasons the model may have performed poorly on this subject (namely, thick skin on the fingertip) are discussed in Discussion, and additional analysis of the model while excluding this subjects' results can be viewed in the Supplementary Materials.

Bland-Altman analysis (bottom two rows of plots in Fig. \ref{reg-ba}) demonstrates the performance of the CNN relative to the transfer standard pulse oximeter in LOOCV.
The SpO\textsubscript{2} values predicted by the learned model near the Limits of Agreement (LOA) reported in previous studies of clinical and non-clinical pulse oximeters, while evaluating on a wider range of SpO\textsubscript{2} levels \cite{luks2020pulse,kelly2001accurate,munoz2008accuracy,lipnick2016accuracy}.  Considering that the ground truth measurements from pulse oximeters exhibit similar variance to these results, this indicates that the model has learned features in the PPG signal that are common across subjects and the model is not simply mean-tracking.
We can also see that for Subjects 1, 3, and 6, there is a negative trend in predictions and the mean difference is above the limits of agreement for some ground truth values in the range 65\% - 80\% SpO\textsubscript{2}. This tells us that the model shows a pattern of consistently over-predicting on SpO\textsubscript{2} samples below 80\%. Notably, without the varied FiO\textsubscript{2} study, we would not have been able to observe the model performance below 85\%, and no prior work has demonstrated that smartphone-based sensing systems may perform poorly at this level. To better understand differences between data ranges, we explore training and evaluating on subsets of our dataset in a data ablation study in the following section.

\label{sec:Data ablation}
\paragraph*{Data ablation}

\begin{figure}[!ht] 
\includegraphics[width=16cm]{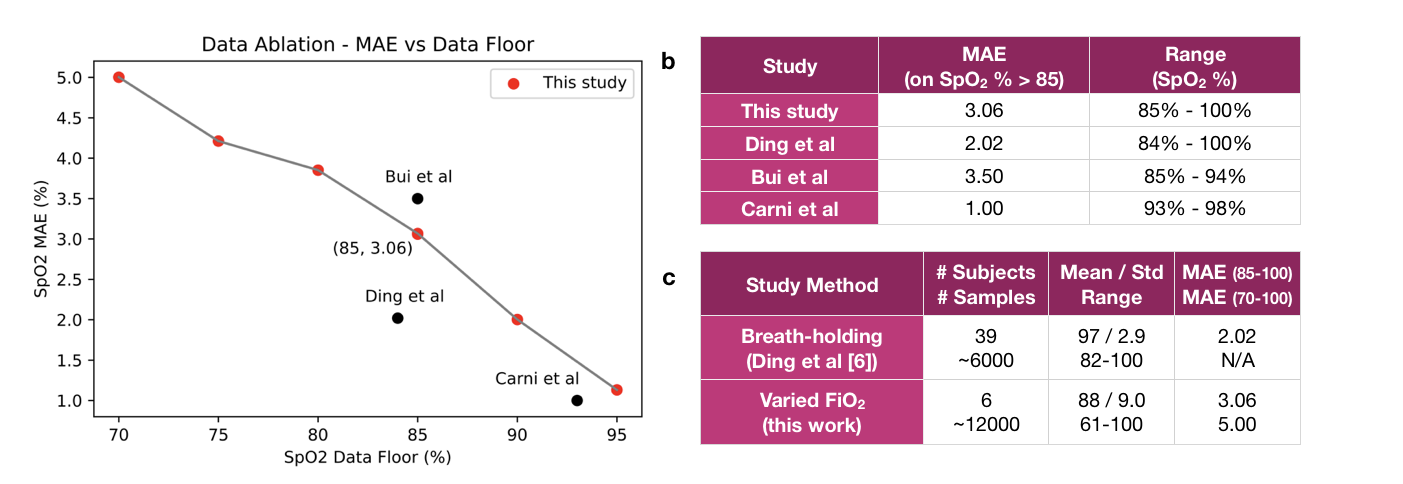}
\caption{\textbf{Data ablation study.} With a data ablation study, we show that our model would have performed better if validated against breath-holding data, which typically does not include data below 80\% SpO\textsubscript{2}. \textbf{a} Mean average error (MAE) of latest works in smartphone-based SpO\textsubscript{2} sensing that perform on datasets with SpO\textsubscript{2} values in the range of 85\% to 100\% \textbf{b} When the range of the data in our work is reduced to a similar level, we achieve comparable accuracy to prior work.  Note that Bui et al used attachments on the smartphone to enhance the photoplethysmographic signal for inference while Ding and the present work use an unmodified smartphone camera \cite{ding2018measuring,bui2020smartphone,carni2016setting}. \textbf{c} Sample statistics and MAE results for this varied FiO\textsubscript{2} study are compared to a recent breath-holding study using smartphone cameras and deep learning \cite{ding2018measuring}.}
\label{abl}
\end{figure}

To understand how the accuracy of our model compares to previously published smartphone-based pulse oximetry systems, we study how excluding subsets of the dataset affects the accuracy. Due to the larger range evaluated in this study compared to prior studies, the overall MAE is not as low as prior studies.  However, a data ablation study reveals that, as lower subsets of the data are removed, the accuracy of our model nears that of other published work. Notably, none of these proof-of-concept works were evaluated on data where a statistically significant portion of the SpO\textsubscript{2} evaluation data was below 85\%, whereas in our varied FiO\textsubscript{2} dataset, the minimum SpO\textsubscript{2} value included is 70\% and the mean of all ground truth SpO\textsubscript{2} levels is 87.1\% (See Fig. \ref{STM-summ}c).

We train and evaluate our machine learning models against a similar dataset to these proof-of-concept works using a data ablation technique. We first subsample our dataset so that we only include samples with ground truth SpO\textsubscript{2} above a floor threshold. We then retrain and evaluate our models to calculate a sub-sampled MAE. Varying across possible thresholds, we observe a negative linear correlation between the minimum SpO\textsubscript{2} value included and the resultant mean absolute error, as can be seen in Fig. \ref{abl}a.  That is, as we reduce the range of SpO\textsubscript{2} values in our training and 
testing dataset, our models perform more accurately. To directly compare to the performance of prior work from Ding et al. and Bui et al. (shown on Fig. \ref{abl}b), we set a SpO\textsubscript{2} threshold of 85\%. While Ding et al. report a range of 73\%-100\%, their dataset shows that only .6\% of all samples are below 85\%, so we report this as a practical floor of 85\% for comparison purposes.
At a floor SpO\textsubscript{2} value of 85\%, our model performs nearly as well as prior work with a mean absolute error of $3.06$\%. 
With this analysis, we can be confident that our techniques are at least as reliable as prior works, and likely benefit from the larger range of training examples.

\label{sec:Classification}
\paragraph*{Classification of hypoxemia}
Rather than simply inferring an estimate for SpO\textsubscript{2}, a smartphone-based tool could be valuable for detecting low SpO\textsubscript{2}, indicating whether or not further medical attention is needed.  To explore the potential of using an unmodified smartphone camera oximeter system as a screening tool for hypoxemia, we calculated the classification accuracy of our model in providing an indication of whether an individual has an SpO\textsubscript{2} level below three different thresholds: 95\%, 90\%, and 85\%. A pulse oximetry value below 90\% SpO\textsubscript{2} is a common threshold used to indicate the need for medical attention \cite{jones2003feasibility}, but other thresholds could be valuable clinically.
Thus, we evaluate the ability of our system to classify samples from our test set by thresholding the regression result from our CNN at different decision boundaries and comparing it to whether the ground truth pulse oximeter simultaneously reports less than the threshold value. We compute sensitivity (true positive rate) and specificity (true negative rate) across all combinations of LOOCV to compute an average result. This experiment simulates the scenario where a smartphone screens a subject it has never seen before, as the model  was trained only on 4 other subjects from the dataset.

\begin{figure}
\includegraphics[width=16cm]{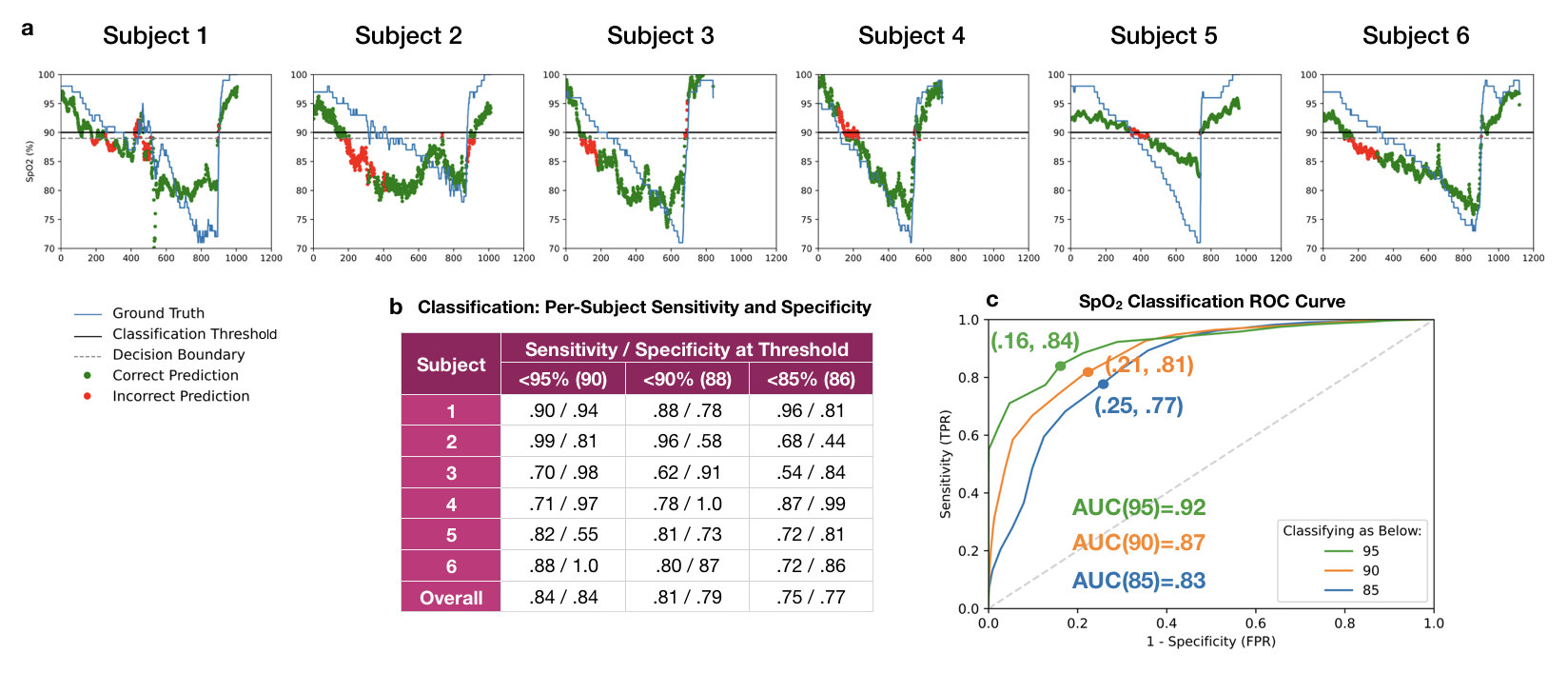}
\caption{\textbf{Classification results for the system.}
\textbf{a} Classifications 
overlaid on ground truth for each subject with a 90\% classification threshold and 88\% decision boundary. 
\textbf{b} Summary statistics for classification across subjects shows that classification performed better on certain patients, and overall achieved a 81\% sensitivity and 79\% specificity rate at sensing whether a subject fell below a 90 \% SpO\textsubscript{2} level
\textbf{c} ROC curves for the classification of low SpO\textsubscript{2}, produced by thresholding the regression model. Classification accuracy decreases as the classification goal is shifted lower, from 95\% to 90\% to 85\%. The classification decision boundary was varied to produce curves for all 3 classification goals, with each point plotted as the average test classification False Positive Rate and True Positive Rate for all LOOCV combinations. The points that are labeled on each curve are those closest to (0,1) for each classification threshold.  The Area Under the Curve (AUC) is .87 for the 90\% threshold SpO\textsubscript{2} level classification.
}
\label{cls}
\end{figure}

The results of this classification analysis can be seen in Fig \ref{cls}. For classifying SpO\textsubscript{2}$<$90\%,
on average across all 6 test subjects, our model attains a sensitivity of 81\% for correctly classifying the positive samples in our dataset of suspected hypoxemia,
while maintaining a specificity of 79\%.
For classifying a subject as below SpO\textsubscript{2}$<$95\%, accuracy increases to sensitivity of 84\% and specificity of 84\%.
Not all combinations of test and train subjects displayed the same level of accuracy. In order to visualize classification accuracy across our entire dataset, we varied the classification decision boundary for three clinically relevant classification thresholds (85\%, 90\%, and 95\%) and averaged the results across all 6 combinations of LOOCV.  The results of varying the decision boundary are plotted on the ROC curve in Fig. \ref{cls}c.  For the SpO\textsubscript{2}$<$90\% classification threshold, the highest accuracy (defined as the closest point to (0,1) on the ROC curve) occurred when the classification decision boundary was set to 88\% SpO\textsubscript{2}.  For clinical value, it may be preferable to choose a threshold that prioritizes sensitivity over specificity, particularly for home settings, so that individuals with low SpO\textsubscript{2} cases would be identified at the expense of over-diagnosis.
For example, choosing a decision boundary of 90\% on the regression result for the SpO\textsubscript{2}$<$90\% classification task enables greater than 92\% accuracy at identifying positive cases (ground truth $<90\%$ SpO\textsubscript{2}), while resulting in 35\% false positives (sensitivity of 92\% and specificity of 65\%).

Classification on individual subjects can be seen in Figure \ref{cls}a. 
The model achieved the best performance on Subject 4, with a sensitivity=88\% and specificity=78\%,
reporting correctly 88\% of the time when the subject had a dangerous SpO\textsubscript{2} level. Subject 5 displayed the lowest sensitivity to specificity tradeoff of 81\% to 73\%. Again, it was noted that the subject had significantly thickened skin on their fingers. Even though the regression for this test subject produces a $MAE=8.56\%$, the classification result indicates that the tool could still be helpful in determining whether or not the user should seek medical attention.

\section*{Discussion}

\paragraph*{Potential for the smartphone as a screening tool}
The classification results from this study indicate a direction to consider for enabling more accessible screening for hypoxemia via unmodified smartphones. Considering the unique positioning of smartphones in the pockets of billions of people worldwide, it would be useful to not only reproduce the function of a pulse oximeter in software, but also to provide an initial screen for clinically significant low SpO\textsubscript{2} levels. The COVID-19 pandemic highlighted this need for an affordable remote oxygen desaturation detection tool that can be accurately and safely used for initial screening and monitoring, informing users whether or not they should seek expert medical attention. This potential is important to consider, as software applications are already being used in this manner even when those applications have not cleared the FDA regulatory requirements \cite{jordan2020utility,digidoc2013plsox}.
Our system is the first unmodified smartphone camera sensor to report accuracy at levels below 85\% SpO\textsubscript{2}, and it achieved relatively high sensitivity (81\%) and specificity (79\%) when classifying subjects with SpO\textsubscript{2} below 90\%. 

\begin{figure}[t!]
    \includegraphics[width=16cm]{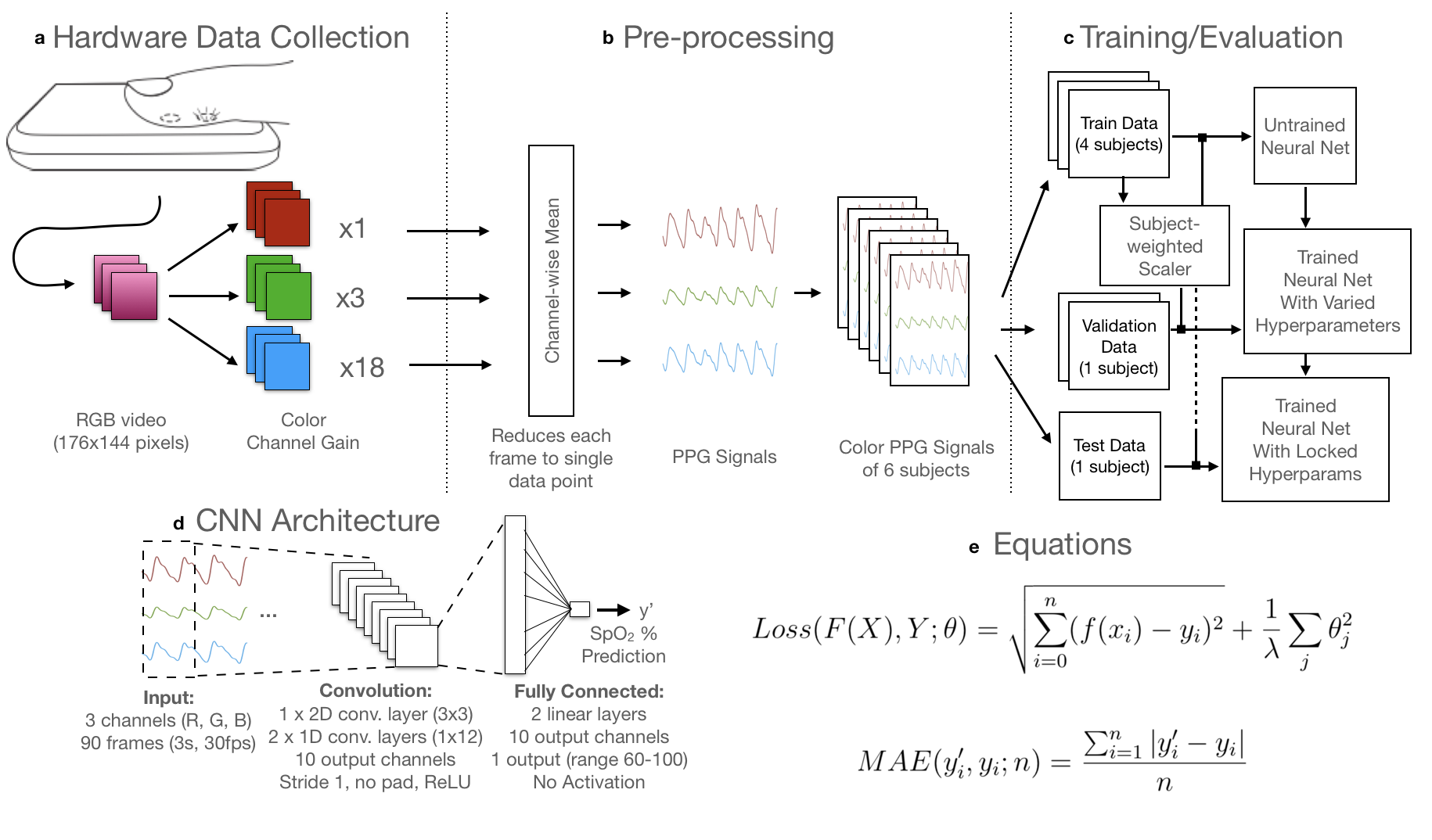}
    \caption{\textbf{Signal extraction and deep learning pipeline}
    \textbf{a} PPG signal extraction occurs after collecting video data from the smartphone camera, applying empirically determined per-channel gains to ensure that each channel is within a usable range (no clipping or saturating). Gains for the R, G, and B channels were empirically determined and held constant throughout all subjects to avoid clipping or biasing towards one channel.
    \textbf{b} Pre-processing of the data computes the PPG signal for each channel as the average pixel value across each frame over time our system. The mean of each channel value across each frame was used as the input for the models.
    \textbf{c} Training and evaluation was performed using Leave-One-Out Cross-Validation (LOOCV) by using 4 subjects' data as the training set, one subject's data as the validation set for optimizing the model, and then evaluating the trained model on one test subject.
    \textbf{d} The deep learning model is constructed of 3 convolutional layers and 2 linear layers operating on the input of 3 seconds of RGB video data (90 frames for 3s at 30fps).  The output is a prediction of the current blood-oxygen saturation (SpO\textsubscript{2} \%) of the individual, which was evaluated using Mean Average Error (MAE) compared to the ground truth standalone pulse oximeter reading.
    \textbf{e} Equations for Loss and MAE that were used in training and evaluating the model.}
    \label{dl}
\end{figure}

\paragraph*{Deep learning}
This SpO\textsubscript{2} prediction pipeline, including smartphone hardware, custom software application, data processing, deep learning and evaluation, is summarized in Fig. \ref{dl}.  Overall, CNN modeling worked well on this input data, learning a function that approximates the data in a non-linear fashion.
This level of performance on a relatively small test subject sample (n=6 subjects with s=12108 total samples) indicates that the model accuracy could increase if more training samples were gathered from further varied FiO\textsubscript{2} experiments, representing a larger range of potential users of the system.

Signal processing and more advanced fine-tuning may improve model performance further, as well as simply gathering more data.
PPG signal noise originating from the wide-band light source and multi-purpose camera sensor of a smartphone likely reduce accuracy. Both prior work and physiological intuition support this theory. Bui et al. suggests that data collected with an unfiltered light source results in a noisy signal that should be modeled with a nonlinear function approximator \cite{bui2020smartphone}. Ding et al. filter the PPG signal to remove noise sources, such as user movement, before feeding the signal to their CNN \cite{ding2018measuring}. While we do not filter our data prior to the CNN, our choice to use a relatively simple neural network architecture with few parameters can be interpreted as a form of regularization that reduces the likelihood of over-fitting. These observations suggest that strong regularization or filtering is important for models predicting SpO\textsubscript{2} from data collected with a smartphone light source and camera.
Overall, though, work in computer vision\cite{Deng2009ImageNetAL} and natural language processing \cite{Brown2020LanguageMA} have shown the effectiveness of collecting higher quality datasets for machine learning. To build a robust model for smartphone-based SpO\textsubscript{2} sensing, a large and broadly collected dataset may be the best way to prevent signal noise from adversely affecting predictions.

In addition, classification machine learning models may improve the classification results.  The only type of machine learning model we tested were in the category of regression algorithms.  Our classification study involved thresholding the result of the regressor.  While that method worked relatively well, producing an AUC of .87 for 90\% classification, we believe that even better classification could be achieved via a classifier model.  Improved classification accuracy would result in enhanced utility as a screening tool, improving the rate at which the sensing system correctly reveals information that allows individuals to seek further care.

\paragraph*{Camera color channel settings}
For this study, camera settings were locked during data gathering by presetting auto-balancing and manually enhancing color gain, which are unique steps in our data collection system relative to prior works in this area.
Camera image capture is variably exposed based on three factors: exposure time, sensor sensitivity, and aperture. For RGB cameras used in smartphones, all three color channels typically use the same exposure time and aperture settings. Both oxygenated and deoxygenated hemoglobin have a significantly higher absorption coefficient in the blue and green wavelengths than for the red wavelengths by about two orders of magnitude. Thus, it would not be possible to measure all three wavelengths simultaneously under the same exposure. If the hardware sensor's sensitivity to a particular color is too high or too low, pixel values for that color may clip by recording the minimum or maximum value of 0 or 255. Because phones use an 8-bit precision scheme for storing pixel data, the pixels will all be rounded to 0 and small changes in that color will be lost. In our application, red is the most dominant color, and prior work has shown that with the use of white balance presets for incandescent light, the tones between blue and green can be amplified \cite{karlen2013detection}. 
	
\begin{figure}
    \centering
    \includegraphics[width=12cm]{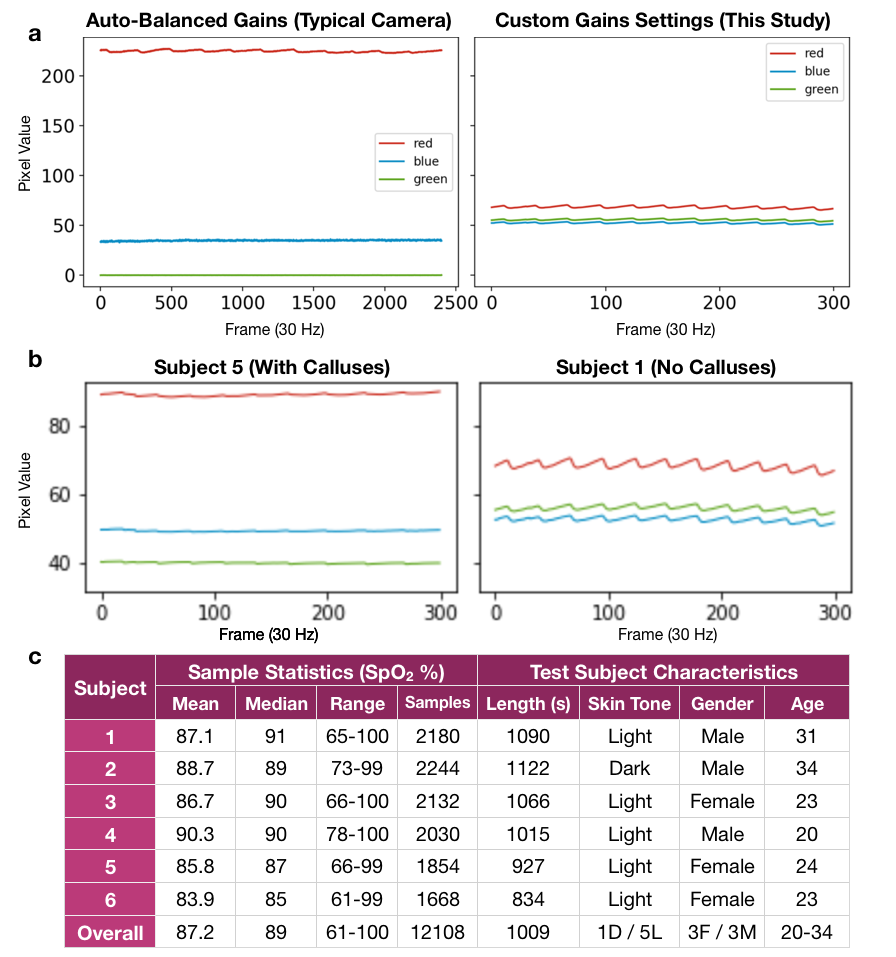}
    \caption{\textbf{Analysis of collected data.} Visualization of PPG data, derived from smartphone videos, reveal the effects of camera gains settings and skin tissue differences on the input signal for our deep learning model. \textbf{a} PPG signal using auto-balance \cite{ding2018measuring} vs custom empirically determined gain settings (this study). 
    In the left image, the green channel is clipped so that the dynamic range becomes so low that the AC variation in the signal cannot be observed. In the right image, the pulsation is visible in all three channels. This shows how standard smartphone camera settings, designed for photography, can reduce the information available to smartphone-based systems for accurate SpO\textsubscript{2} sensing.
    \textbf{b} Skin tissue aberrations (such as calluses seen in Subject 1's fingers) can affect the quality of data available for SpO\textsubscript{2} sensing.  At left, the raw data in the red, blue, and green channels for Subject 5 are dampened and the AC portion of the signal cannot be observed at a resolution of 300 frames. At right, the AC portion can be clearly seen for Subject 2 at the same resolution.
    This abnormality is likely due to Subject 5's callused tissue on the fingers.
    \textbf{c} Subject breakdown for the FiO\textsubscript{2} study and ground truth data statistics (in SpO\textsubscript{2} \%) for each subject. The average difference between mean and median for each subject is 1.58, showing minimal skew. The average length of each subject's test run is about 16 minutes.}
    \label{RGB-ana}
\end{figure}

Software advancements in smartphone image processing pipelines now provide more independent control of each color channel's exposure through independent per-channel amplifier gain settings. By having control of independent amplifier gain settings, we can balance the exposure settings to amplify the blue and green channels. Different operating systems allow for a different granularity in the gain control settings. Our work was enabled with the Android Camera2 API, which provides access to manual setting of sensitivity, exposure, and individual color gains. Therefore, to ensure that the blue and green signals are not lost, we empirically determined and assigned a fixed color gain in our application, ensuring that a usable signal is recorded by the camera for all 3 channels. We empirically determined and used the gains (1,3,18) for R, G, and B (See Fig \ref{dl}a). After we set the color channel, we determined through the same study the use of 1.2ms for exposure time and a sensor sensitivity of 300 ISO performed well in evenly exposing R,G, and B color channel PPG signals at the middle of the 0-255 value range. The impact of this on the RGB signal is shown in Figure \ref{RGB-ana}a.
With auto-balancing, the red PPG clips fall at the top of the 0-255 range of RGB lumen values, while the green channel falls to close to 0. In comparison, using custom hardware gain settings in this study, controlled through the Android Camera2 API, allowed all three color channel PPGs to be well-represented in the 8 bit range and display differentiable AC signal amplitude. 

\paragraph*{Skin tissue aberrations}
We see particularly aberrant performance
on subject 5 with MAE=8.56. We suspect this is due to exacerbated tissue noise on the subject's fingers from thickened skin, which is not represented in the rest of the training data. This subject was noted to be the only subject in the study with noticeable calluses on their fingertips, and the subject indicated this was due to sports. We investigate the data obtained from this subject more closely in Fig. \ref{RGB-ana}b and observe that the PPG signals for subject 5 show nearly 50\% dampened oscillations (AC signal component), quantified by a standard deviation of 3.44 compared to 6.86, and 50\% higher average value (DC signal component), quantified by a mean of 84.5 compared to 51.5, relative to other subjects.
We hypothesize that these abnormal features are a result of the calluses. Specifically, an abnormally thick layer of tissue on the finger would absorb more light in the blue and green spectra. Because our device's sensor has fixed sensitivity, the abnormally attenuated light in the blue and green spectra results in poor measurement of the pulsatile blood and altered spread in color channel values.
With a small training set of 4 subjects including no other examples of subjects with fingertip calluses, the model cannot learn to account for these tissue differences. We anticipate the model could learn to account for tissue abnormalities if trained on more subjects, if adaptive gain settings were employed to gather data that ensured a similar oscillation amplitude in the AC signal for the input data collected by the smartphone.

\paragraph*{Limitations}
From this limited dataset, we are unable to make definitive conclusions regarding the effect of skin tone or gender on smartphone pulse oximetry. Our test subjects included 1 subject with a dark skin tone (subject 2 identified as African-American) and 5 subjects with a light skin tone (all other subjects identified as Caucasian), as seen in Fig. \ref{RGB-ana}c. Our model does not appear to perform differently based on skin tone with this limited dataset, as the results for subject 2 fell in a similar range as other subjects, as seen in Fig. \ref{reg-ba}. However, it has been shown that standalone pulse oximeters, such as the one used as the ground truth in our dataset, can produce decreased accuracy on patients with darker skin tones \cite{feiner2007dark}.  Based on our study, we do not claim any findings that our model works better or worse based on skin tone, but that should be evaluated in future studies.  Our model also does not perform differently on either side of our 3:3 female:male gender split. Analyzing performance of our model to users of different skin tones and genders is important, but will require further work to understand.

This work on smartphone camera oximetry locks exposure settings rather than allowing the phone to auto-balance. Other research published around smartphone camera oximetry have relied on phone auto-balancing features and out of the box white-balancing algorithms. We made the design choice to appropriate the phone sensors to act much more like a simple sensor system that doesn't automatically adjust in software; however, when we examine the performance that has been achieved in our data ablation study, it may be the case that both methods are sound approaches. Our study does not help to elucidate whether auto-balancing would work well at lower ranges; therefore, it would be useful to perform a similar study for auto-balancing based camera oximetry systems.

In this study, we did not analyze signal filtering for motion artifacts. While other recent works employ signal processing methods for removing noisy data points \cite{bui2020smartphone,ding2018measuring}, such as finger slippages or shaking (as is common in breath-holding experiments), the controlled design of the experiment allows the finger to stay relatively stable and consistently coupled to the sensor, as compared to breath-holding experiments. However, in any future screening tool, it will be important in real-world use to detect excessive motion, provide feedback to the user to keep still, and discard high motion segments \cite{petterson2007effect}.  Combining our methods with signal processing techniques studied in prior works may improve results in clinical and outpatient use.

\paragraph*{Study Expansion}
Informing directions for future work in this area, we note that our study size was limited due to the cost of running a FiO\textsubscript{2} study, which is approximately \$8,000 for the 6 subject tested. While we can speculate that this prediction task becomes more difficult at lower SpO\textsubscript{2} ranges based on our Bland-Altman statistics in Fig. \ref{abl}, we need to collect more data in the 70\% to 85\% range to better understand why this happens. It is still important to note that this finding of diminished performance at lower than 85\% oxygen saturation is a core contribution of this work as the first published study (to our knowledge) to test smartphone camera oximetry in an induced hypoxemia study on the full range of clinically relevant data. 

Furthermore, the study we presented uses a transfer standard method of validation using optical pulse oximeters as the ground truth gold standard reference. FDA clearance of new pulse oximetry devices requires 
a full human desaturation study, including regular blood draws for co-oximetry validation against an ABG ground truth \cite{us2013pulse}.
Because that test is more expensive ($>$\$20,000) and invasive, it is common to perform a transfer standard pulse oximeter-based study, such as the one performed in this study, during development of a new device prior to a full human desaturation study \cite{clinimark2010pulse}. After more development and validation using this transfer standard method, a full human desaturation study could be warranted.

\paragraph*{Conclusion}
Our results, in this pilot study of 6 subjects, provide a positive indication that a smartphone could be used to assess risk of hypoxemia without the addition of extra hardware. In order to validate and enable this in the future, we would recommend gathering more data with a smartphone in varied FiO\textsubscript{2} studies that induce hypoxemia to increase the training data and the accuracy of the deep learning model. With an improved model, we could set up user studies in which the app is used in conjunction with a standalone pulse oximeter to measure the accuracy of the software-based solution in real-world scenarios.
We would also like to see what others in the community can do with the open-source FiO\textsubscript{2} data that we are providing alongside this paper. More development and testing could allow this tool to become beneficial for low-cost clinical management of individuals with chronic respiratory conditions, such as COPD, as well as acute respiratory diseases like COVID-19.

\section*{Materials and Methods}

\paragraph*{Study Design}
    6 healthy test subjects were recruited and enrolled to participate in a varied FiO\textsubscript{2} study to evaluate the efficacy of using unmodified smartphone cameras in pulse oximetry.
	The varied FiO\textsubscript{2} study was performed using the varied fractional inspired oxygen protocol administered by a clinical validation laboratory, Clinimark, which is a group that performs validation services for medical devices \cite{clinimark2010pulse}. This experiment was approved by the Internal Review Board at Clinimark. Consent for each participant was obtained prior to commencing the test procedure. Six subjects were administered controlled fractional mixtures of medical grade oxygen-nitrogen in a controlled hospital setting for 14-19 minutes. The subjects rested comfortably in a reclined position while the gas mixture was given to induce hypoxemia in a stair-stepped manner. During this time, the subjects' fingers were instrumented with multiple transmittance pulse oximeter clips and two smartphone devices, with the smartphone device on the index finger of each hand. The ground truth data was recorded using multiple purpose-built pulse oximeters, including a tight-tolerance transfer standard pulse oximeter, the Masimo Radical-7 \cite{masimo7radical}. Subject characteristics and data statistics can be seen in Fig. \ref{RGB-ana}c. Subject observations were recorded, including the observation that one subject, Subject 5 in the analysis, had particularly callused hands.
	
\paragraph*{Smartphone Device Configuration and Setup}
	We collected camera oximetry data with a Google Nexus 6P, recording video at 30 frames per second in a custom video capture application developed in Java using Android Studio. The device was specifically configured so that camera exposure settings in the camera hardware did not change throughout the entire study. Color gains were set to 1x for the red channel, 3x for the green channel, and 18x for the blue channel. These gains were chosen empirically by manually analyzing the impact of gain value adjustments on 20 healthy individuals to find gain values that avoided data loss due to compression and obtained optimal signal quality (see Fig. \ref{RGB-ana}a). During the varied FiO\textsubscript{2} study, because the device could overheat from recording continuous video with flash enabled for more than 1 minute, we placed clay ice packs around the device to keep its temperature down for the 14-19 minute duration of the study. The ice packs were placed strategically to avoid contact with the hand.

\paragraph*{Data pre-processing}
For each hand on each subject, we recorded an ordered list of $n$ RGB image frames, each with $176\times144$ pixels. To obtain a PPG signal, we computed the mean pixel value for each color channel and obtained a $3\times n$-shaped matrix of values. 
Each hand of each subject is treated as a unique subject in the display of results. 
We divide the data into samples for each 1-second (30 frames) window, combining the 3 seconds (90 frames) of sample RGB data centered on 1 ground truth SpO\textsubscript{2} reading as one sample. This provides over 8000 training examples (4 subjects) to our models, with about 2000 samples (1 subject) held out for both the cross-validation and test set for each configuration of LOOCV. Samples under 70\% SpO\textsubscript{2} are removed prior to training and validation due to the sparsity of samples in that range.

\paragraph*{Convolutional neural network}
We applied a CNN machine learning model, detailed in Fig. \ref{dl}. 
We designed and trained a network with three convolutional layers 
followed by two fully connected layers.
For the first convolution, we treat the RGB channel components of our signals as a second dimension and use kernel sizes of $3\times3$ with no padding. We normalize and standardize both training and validation datasets based on a weighted channel-wise mean and standard deviation of the training dataset, where the weights are scaled by the length each subject's data collection. 
The model is trained using the Adam optimizer with a learning rate of 0.00001 (with a rate decay by 0.1 after 80 epochs) and an L2 regularization of strength 0.1. We optimize Mean Squared Error (MSE) as our loss function and report the accuracy of the results by computing the MAE (Fig \ref{dl}e). The model is built and trained using the PyTorch library.

\paragraph*{Statistical analysis}
We identified and evaluated two potential usage scenarios for a software-based oximetry solution on a standalone smartphone: (1) as a replacement for traditional pulse oximeters by regressing a continuous SpO\textsubscript{2} value, and (2) as an at-home screening tool to inform the need for a follow-up with a physician by classifying regression results as below a particular threshold. 

We explored the first scenario of pulse oximetry measurement by performing a regression analysis, comparing our smartphone measurement to a purpose-built pulse oximeter with error and Bland-Altman metrics. In our performance assessment, we evaluated models using Leave-One-Subject-Out cross validation (LOOCV). Specifically, we trained and tested on six validation splits, with two different subjects (both hands) held out for cross-validation and testing in each split. We visually examined the ground truth distributions of the splits to ensure there was not a heavy imbalance in the dataset. We compared the performance of algorithms using Mean Absolute Error.

We explored the second scenario of hypoxemia screening by performing a classification analysis, thresholding the ground truth recordings below 3 different SpO\textsubscript{2} levels (95\%, 90\%, and 85\%) and comparing it to our thresholded regression result. We examined the true positive (sensitivity) and true negative (specificity) rates at different screening decision boundaries (95\%, 90\%, and 85\%) to illustrate the potential performance of the system for use in hypoxemia screening. To interrogate the potential to adjust this decision boundary to bias towards sensitivity or specificity, we varied the decision boundary across the range of 70\%-100\% and plotted ROC curves for each subject using LOOCV.

\paragraph*{Data Availability}
We provide the data from the varied FiO\textsubscript{2} study in open source format to the community to allow others to build upon this work.

\section*{List of Supplementary Materials}
\begin{enumerate}
    \item Fig. S1 Regression Results After Removing 1 Subject
    \item Fig S2. Classification Results After Removing 1 Subject
    \item Data file S1. Zip file of raw camera oximetry data
\end{enumerate}

\section*{References and Notes}
\bibliography{scifile}

\bibliographystyle{Science}

\smallskip

\textbf{Acknowledgments:}
The authors thank Clinimark for conducting the study.
\textbf{Funding:}
University of Washington gift funding supported the study.
\textbf{Author contributions:}
EJW and SNP designed the study.  EJW built the smartphone application and collected data.  JSH and VV built the machine learning model.  ECL and XD collected comparison data and advised on study design.  JSH, VV, and EJW wrote the manuscript.  All coauthors edited the manuscript.
\textbf{Competing interests:} 
JSH, SNP, EJW, and VV are inventors of US patent application 17/164,745 covering systems and methods for SpO\textsubscript{2} classification using smartphones.
\textbf{Data and materials availability:}
All data associated with this study are available in the main text or the supplementary materials.

\newpage
\section*{Supplementary Materials}

\textbf{Fig. S1 Regression Results After Removing 1 Subject}

\smallskip

\begin{figure}[h]
\includegraphics[width=16cm]{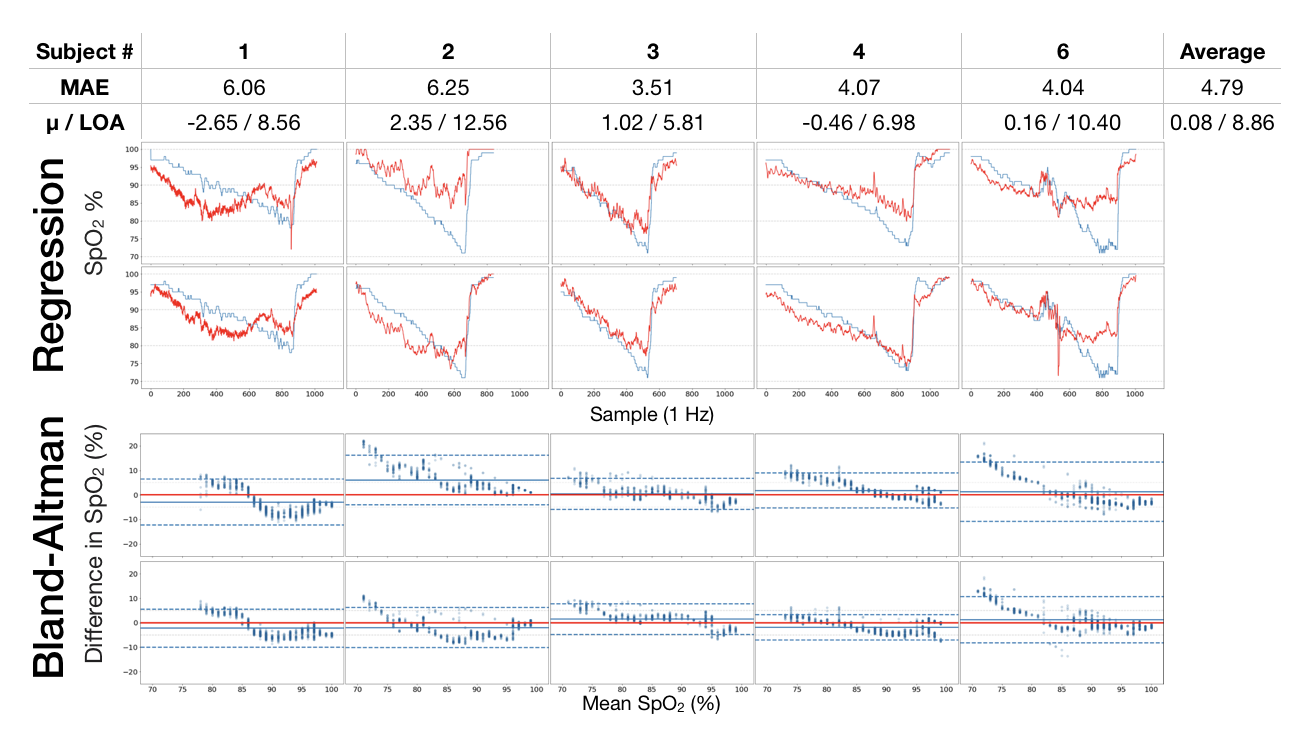}
\caption{\textbf{Regression results with one subject's data removed from the model, displayed as direct performance analysis and Bland-Altman comparison.} Mean Average Error (MAE) averages to 4.79 over 5 subjects after subject 5 is removed from the model training and validation.  The average difference ($\mu$) and limits of agreement (LOA) average to 0.08 and 8.86, which compare favorably to standalone pulse oximetry devices \cite{kelly2001accurate}. \textbf{Table:} Mean Average Error (MAE) and Bland-Altman statistics for CNN evaluation by LOOCV for these subjects (n=5). \textbf{Regression:} Plots of direct performance analysis of regression results.  Model predictions (in blue) and ground truth readings (in red) for the 6 subjects in the FiO\textsubscript{2} study plotted against time of study.  Left hand is on top and right hand is on bottom. \textbf{Bland-Altman:} Bland-Altman plots displaying the spread of predictions against ground truth, revealing that the standalone pulse oximeter and smartphone model perform similarly, and more closely aligned when the one subject with calluses is removed from the analysis.}
\label{reg-ba-5}
\end{figure}

\newpage

\smallskip

\noindent
\textbf{Fig S2. Classification Results After Removing 1 Subject}

\smallskip

\begin{figure}[h]
\includegraphics[width=16cm]{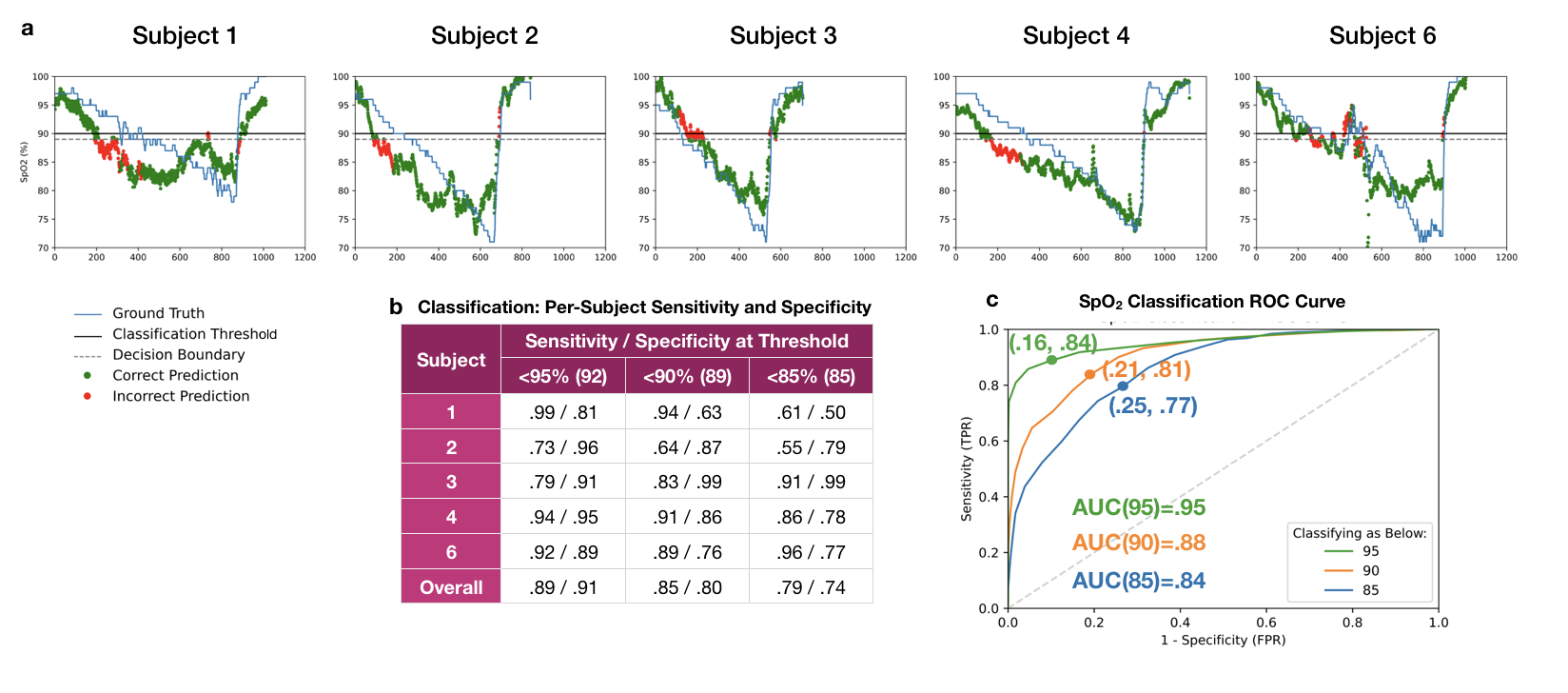}
\caption{\textbf{Classification results for the system with 1 subject's data removed.}
\textbf{a} Classifications overlaid on ground truth for each subject with a 90\% threshold and 89\% decision boundary. 
\textbf{b} Summary statistics for classification across subjects shows that classification performed better on certain patients, and overall achieved a 85\% sensitivity and 80\% specificity rate at sensing whether a subject fell below a 90 \% SpO\textsubscript{2} level
\textbf{c} ROC curves for the classification of low SpO\textsubscript{2}, produced by thresholding the regression model. Classification accuracy decreases as the classification goal is shifted lower, from 95\% down to 85\%. The classification decision boundary was varied to produce curves for all 3 classification goals, with each point plotted as the average test classification false positive rate and true positive rate for all LOOCV combinations. The points that are labeled on each curve are the closest to (0,1).
}
\label{cls-5}
\end{figure}

\smallskip

\noindent
\textbf{Data file S1. Zip file of raw camera oximetry data}

\noindent
A zip file of raw data from the FiO\textsubscript{2} study will be shared after peer review and publication.

\end{document}